\documentclass[conference,a4paper]{IEEEtran}
\usepackage{amsmath,amssymb,booktabs,graphicx}
\usepackage{cite}
\usepackage[hidelinks]{hyperref}
\begin{document}
\bstctlcite{IEEEexample:BSTcontrol}
\title{Cross-Version Compatibility of TESSERA Embeddings Through Lightweight Translation}
\author{
\IEEEauthorblockN{Yutong Yang}
\IEEEauthorblockA{Department of Computer Sciences\\
University of Wisconsin--Madison\\
Madison, WI, USA\\
\texttt{yang844@wisc.edu}}
\and
\IEEEauthorblockN{Zhengpeng Feng\textsuperscript{*}}
\IEEEauthorblockA{Department of Computer Science and Technology\\
University of Cambridge\\
Cambridge, United Kingdom\\
\textsuperscript{*}Corresponding author: \texttt{zf281@cam.ac.uk}}
}
\maketitle
\begin{abstract}
Downstream models often outlive the embedding release on which they were
trained. We test whether lightweight converters let a classifier trained on
one Tessera release use embeddings from another. We fit converters between
v1, v1.1, and v2 on 6,280 co-located triplets from 25 sites in 2024, with
disjoint training, validation, and test sites. On a four-class land-cover
test, a classifier trained on true v2 reaches 0.937 accuracy and 0.844
macro-F1. The same classifier scores 0.061 and 0.050 on raw v1.1, but 0.917
and 0.795 on converted v1.1. Direct v1-to-v2 translation has lower error than
composition through v1.1. Geographic transfer is less reliable. On a separate
cohort of 20 sites, a converter trained on the original sites reaches 0.597
accuracy despite 0.902 cosine similarity to v2, whereas a converter trained
within the new cohort reaches 0.784. Tests with 2021 WorldCover labels and at
24 sampled sites without valid native v2 values extend the evaluation.
Conversion recovers much of the frozen classifier's performance on the tested
sites, but the new-site result shows that embedding similarity cannot stand
in for downstream evaluation.
\end{abstract}
\begin{figure*}[t]
\centering
\includegraphics[width=\textwidth]{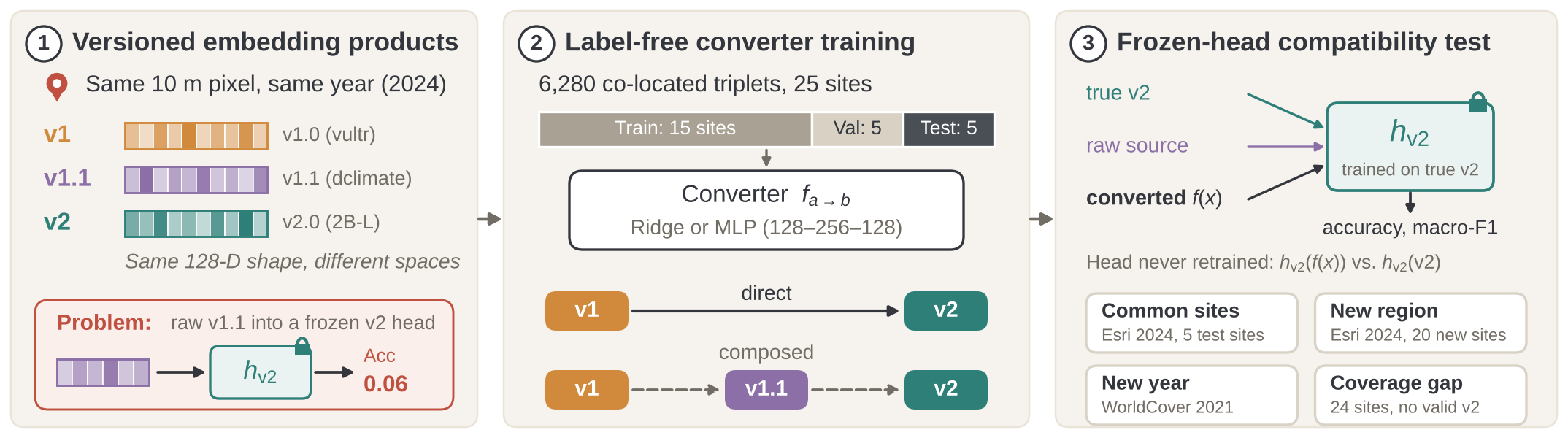}
\caption{We test whether converted embeddings can replace native ones as the
input to a fixed classifier. We align v1, v1.1, and v2 embeddings at the same
locations and year, fit ridge and MLP converters on spatially disjoint
training sites, and compare native, raw source, and converted inputs to a
classifier trained only on true target-release embeddings. The evaluation also
tests converter composition, geographic transfer, a second year, and
locations without valid native v2 values.}
\label{fig:overview}
\end{figure*}

\section{Introduction}
\label{sec:intro}

Embedding products, like software libraries, are released in versions. A new
library version can break programs written against the old interface, and a
new embedding version can break classifiers trained on the old embeddings.
Tessera, which summarizes the Sentinel-1 radar and Sentinel-2 optical time
series of one year as a 128-dimensional vector per 10\,m pixel
\cite{feng2026tessera,feng2026applications}, has had three releases. Version 1
trains a pixel-wise Barlow Twins encoder
\cite{zbontar2021barlow,lisaius2024barlow,feng2026tessera}. Ball et al.\ found
that a species map fitted to one year of v1 embeddings is less accurate when
applied to a different year \cite{ball2026species}, and version 1.1 retrains
the encoder for stability across years \cite{feng2026hpc}. Version 2 distils a
2-billion-parameter teacher into compact students that produce Matryoshka
embeddings \cite{kusupati2022matryoshka,feng2026tesserav2}. Because every
release is 128-dimensional, a classifier trained on one release accepts input
from another and returns predictions without any warning, even though the
coordinates of the two releases have different meanings.

The cost of this incompatibility falls on downstream users. Users of embedding
products such as Tessera and AlphaEarth Foundations
\cite{feng2026tessera,brown2025alphaearth} train their own classifiers, and
Tessera-based classifiers already support forest species mapping
\cite{ball2026species}, crop classification in Senegal
\cite{lisaius2026senegal}, burned-area mapping \cite{knezevic2026wildfire},
and a range of other environmental mapping tasks \cite{feng2026applications}.
Models that forecast surface change in embedding space
\cite{li2026forecasting} depend on the release in the same way. When a new
release appears, each user must either retrain on it, which requires the
original labels, or keep the old release and forgo the improvements. Neither
choice helps where the new release has gaps: at some locations we sampled, the
queried v2 release returns no valid embedding although v1.1 does.

Translation between learned representations has precedent in computer vision.
Lenc and Vedaldi found that the features of one convolutional network can
often be mapped into another by a learned transformation
\cite{lenc2015equivalence}, and such model stitching has since been used to
compare representations \cite{bansal2021stitching}. Backward-compatible
training takes a different route. It constrains a new model during training,
so that its embeddings can be compared directly with those of the old one
\cite{shen2020bct}, which requires compatibility to be planned before a
release. FastFill instead learns a transformation from old to new features
once both models exist, so that only part of a retrieval gallery needs to be
recomputed \cite{jaeckle2023fastfill}. We show that Tessera releases are not
compatible, so they too need a map fitted after release. Earth observation
embedding products have so far been compared release by release on downstream
benchmarks \cite{feng2026tesserav2}, but, to our knowledge, not translated
from one release to another.

We fit lightweight converters (ridge regression and a small multilayer
perceptron) between co-located v1, v1.1, and v2 embeddings \cite{geotessera}
and judge them by \emph{downstream compatibility}: a classifier trained on the
target release is kept fixed while its input is replaced by converted source
embeddings (Fig.~\ref{fig:overview}). A converter can help only to the extent
that the source embedding retains what the classifier needs. We evaluate the
converters with Esri land-cover labels \cite{karra2021esri,esri} at held-out
sites in 2024 and at 20 new sites, with ESA WorldCover labels
\cite{worldcover} in 2021, and at 24 sites where v2 has no valid value. We
make three main findings.
\begin{itemize}
\item Conversion recovers much of the lost accuracy. A classifier trained on
v2 scores 0.937 on true v2, 0.061 on raw v1.1, and 0.917 on converted v1.1.
\item Direct v1-to-v2 translation has lower error than composition through
v1.1.
\item On 20 new sites, a converter with higher cosine similarity to v2 gives
lower classifier accuracy (0.597 versus 0.784), so embedding similarity cannot
stand in for downstream evaluation.
\end{itemize}
Section~\ref{sec:methods} describes the data and converters,
Section~\ref{sec:results} reports these results in full, and
Section~\ref{sec:discussion} discusses what they do and do not establish.

\section{Data and methods}
\label{sec:methods}

\subsection{Aligned releases and site splits}
We pair v1.0/vultr, v1.1/dclimate, and v2.0/2B-L$\sim$beta1 embeddings at the
same coordinates in 2024. The main dataset contains 6,280 triplets with finite
values from 25 sites, and each vector has 128 dimensions. We assign 15 sites
(3,835 samples) to training, five (1,167) to validation, and five (1,278) to
testing, and all three converters use these splits. Holding out whole sites
rather than random points reduces the optimistic bias that spatial
autocorrelation introduces into model evaluation
\cite{roberts2017cv,ploton2020spatial}.

At some checked historical points, previously stored v2 vectors differed from
v2 vectors read afresh from the NPY tiles. We therefore rebuilt the v2 targets
for all 6,280 triplets from the NPY tiles and use the rebuilt vectors in the
primary analysis, keeping the stored targets for sensitivity runs. The source
of the numerical discrepancy remains unresolved.

To test geographic transfer, we assemble 2,560 v1.1/v2 pairs from 20
additional sites in 2024, with no sites or coordinates in common with the main
cohort. Its training, validation, and test sets contain 1,536, 512, and 512
samples across 12, four, and four sites. To test a second year, a 2021 cohort
adds ESA WorldCover labels to v1.1/v2 pairs at 16 sites. Its three-class task
has 979 training, 190 validation, and 457 test samples. Finally, we sample 379
valid v1.1 vectors at 24 selected sites across six continents where the
queried v2 release returns no valid vector, so this pilot has no
same-location v2 reconstruction target.

\subsection{Converters and evaluation}
For source vectors $x_i$ and target vectors $y_i$, we fit a map
$f_{a\rightarrow b}$ and select models by validation MSE. Ridge regression
\cite{hoerl1970ridge} provides a linear baseline, with its regularization
weight chosen from
$\{10^{-6},10^{-4},10^{-2},10^{-1},1,10,10^2,10^3\}$. The MLP has a
$128\rightarrow256\rightarrow128$ architecture with GELU activation
\cite{hendrycks2016gelu} and dropout 0.1. We standardize inputs and targets
with training-set statistics and train with AdamW
\cite{loshchilov2019adamw} (learning rate $10^{-3}$, weight decay $10^{-4}$,
batch size 256), stopping early on validation MSE. We train five fixed seeds
and average their predicted embeddings for downstream evaluation. To test
composition, we compare direct $f_{1\rightarrow2}(x)$ with composed
$f_{1.1\rightarrow2}(f_{1\rightarrow1.1}(x))$ at identical test coordinates.

We measure reconstruction by MSE, mean cosine similarity, global $R^2$, Top-1
retrieval, and nearest-neighbour overlap, where retrieval searches only among
the held-out samples. For each downstream task, we train a standardized,
class-weighted logistic regression head on true target-release embeddings and
select its regularization by validation macro-F1. The fitted head then stays
fixed while raw or converted source vectors are substituted at test time. We
report accuracy and macro-F1 because class supports are uneven. Paired
intervals come from a bootstrap \cite{efron1993bootstrap} that resamples
held-out sites rather than points, because nearby points are not independent.

\section{Results}
\label{sec:results}

\subsection{Translation and composition on common sites}
\begin{figure}[t]
\centering
\includegraphics[width=\columnwidth]{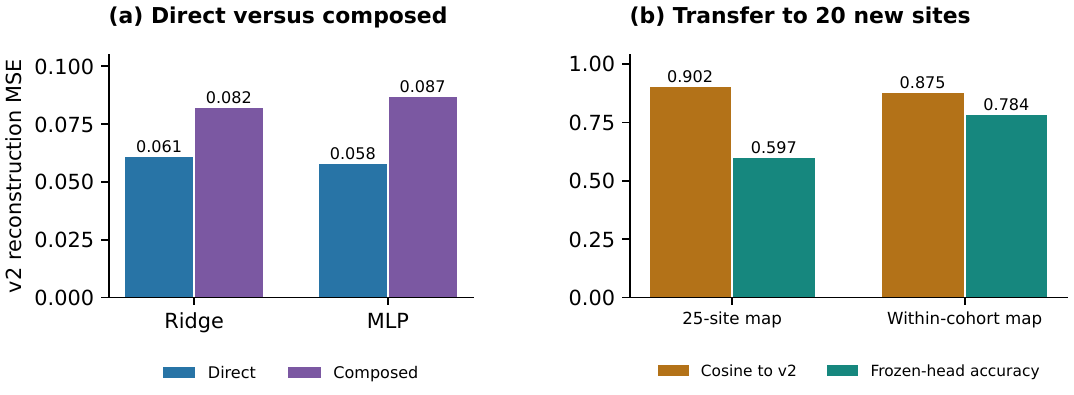}
\caption{Composition adds error, and cosine similarity does not predict
classifier accuracy. (a) Direct v1-to-v2 conversion has lower v2
reconstruction MSE than composition through v1.1 for both model families on
the common 2024 test sites. (b) On the separate 20-site cohort, the converter
trained on the 15 training sites of the original 25-site cohort has higher cosine similarity to v2 but lower
accuracy through the frozen true-v2 head than the converter trained within the
new cohort. Bar heights in (b) represent different metrics.}
\label{fig:contrasts}
\end{figure}

Table~\ref{tab:translation} summarizes translation on the 1,278 common-site
test triplets. The v1.1$\rightarrow$v2 MLP reaches 0.962 mean cosine
similarity, but its within-test-set Top-1 retrieval is 0.280. Direct
v1$\rightarrow$v2 translation has lower MSE than composition through v1.1,
both for ridge (0.06060 versus 0.08197) and for the MLP (0.05755 versus
0.08660), as Fig.~\ref{fig:contrasts}a shows. Note that v1$\rightarrow$v1.1 is a map to a different target space (MSE 0.39479), so its error cannot be compared directly with v2-target MSEs. The site-bootstrap intervals for
the composed-minus-direct MSE are $+0.02137$ [0.01016, 0.03896] and
$+0.02866$ [0.01171, 0.05443], respectively, so composition introduces
measurable error on these test sites.

\begin{table}[t]
\centering
\caption{Direct v1$\rightarrow$v2 translation reconstructs v2 most closely.
MLP translation on the common 2024 held-out sites, as means across five seeds,
with v2 targets read from NPY tiles.}
\label{tab:translation}
\setlength{\tabcolsep}{3pt}
\begin{tabular}{lrrrr}
\toprule
Converter & MSE$\downarrow$ & Cosine$\uparrow$ & $R^2\uparrow$ & Top-1$\uparrow$ \\
\midrule
v1$\rightarrow$v1.1 & .39479 & .88677 & .58128 & .28764 \\
v1.1$\rightarrow$v2 & .07535 & .96182 & .68535 & .27966 \\
v1$\rightarrow$v2 & .05755 & .97109 & .75967 & .52504 \\
v1$\rightarrow$v1.1$\rightarrow$v2 & .08660 & --- & --- & --- \\
\bottomrule
\end{tabular}
\end{table}

\subsection{Frozen-head compatibility and geographic transfer}
Table~\ref{tab:heads} reports the frozen-head results. The main four-class
Esri 2024 test \cite{esri} contains 1,238 points at five held-out sites, on
which the true-v2 head scores 0.937 accuracy and 0.844 macro-F1. The same head
scores 0.061 and 0.050 on raw v1.1, but 0.917 and 0.795 on translated v1.1. A
separate head trained on true v1.1 behaves similarly: its accuracy is 0.919 on
native inputs, 0.133 on raw v1, and 0.903 on converted v1. In each comparison
the head stays fixed and only its input changes.

The 20-site cohort tests transfer beyond the original sites. Its four-class
Esri test has 509 supported labels at four held-out sites. Through the frozen
new-cohort v2 head, a v1.1$\rightarrow$v2 MLP trained on the 15 training sites of the original 25-site cohort
scores 0.597 accuracy, whereas an MLP trained and validated within the 20-site
cohort scores 0.784. Yet on these points the original converter has the lower
reconstruction MSE (0.18192 versus 0.23111) and the higher cosine similarity
(0.90194 versus 0.87519), as Fig.~\ref{fig:contrasts}b shows. Average vector
error therefore fails to predict which converter works better with this head.
The two models also differ in training locations, sample counts, and
selection, so we cannot attribute the gap to a single cause. Note that class 2
accounts for 343 of the 509 test points, including all points at one site.

To test a second year, we apply the 2024-trained converter to the three-class
ESA WorldCover 2021 task \cite{worldcover}. On 457 points at four held-out
sites, the true-v2 head scores 0.867 accuracy and 0.864 macro-F1. Converted
v1.1 scores 0.783 and 0.807, and raw v1.1 scores 0.254 and 0.150. The
site-bootstrap 95\% interval for the converted-minus-true-v2 accuracy
difference is [$-0.222$, 0.037]. The interval is wide because there are only
four test sites and they vary markedly.

\begin{table}[t]
\centering
\caption{Converted inputs recover much of the accuracy that raw inputs lose.
Inputs to frozen true-v2 land-cover heads. The main Esri results use the
NPY-target MLP, and the 20-site ``within'' model is trained within the
separate cohort.}
\label{tab:heads}
\setlength{\tabcolsep}{3pt}
\begin{tabular}{llrr}
\toprule
Task & Input & Accuracy & Macro-F1 \\
\midrule
Esri main, 2024 & True v2 & .937 & .844 \\
& Raw v1.1 & .061 & .050 \\
& Converted v1.1 & .917 & .795 \\
& Direct converted v1 & .923 & .808 \\
& Composed converted v1 & .899 & .749 \\
\midrule
Esri new sites, 2024 & True v2 & .770 & .758 \\
& Raw v1.1 & .104 & .068 \\
& Original-cohort converter & .597 & .646 \\
& Within-cohort converter & .784 & .767 \\
\midrule
WorldCover, 2021 & True v2 & .867 & .864 \\
& Raw v1.1 & .254 & .150 \\
& Converted v1.1 & .783 & .807 \\
\bottomrule
\end{tabular}
\end{table}

\subsection{Coverage-gap pilot and sensitivity}
At the sampled points of the 24 selected sites, the queried v2 release has no
valid vector. Of the 379 valid v1.1 vectors there, 299 have labels supported by
the frozen head, and we exclude the other 80. On the supported subset, an
NPY-target v1.1$\rightarrow$v2 MLP paired with an NPY-target v2 head scores
0.712 accuracy and 0.811 macro-F1, whereas a separately trained native-v1.1
system scores 0.538 and 0.519. Note that this compares two complete systems
whose heads were trained on different releases. Without same-location true v2,
the pilot cannot measure agreement with native-v2 predictions. The selected
sites probe coverage gaps and do not represent all global land.

Rebuilding the common-site targets from NPY tiles changes individual results,
although the main compatibility pattern remains. On the new-site task, using
NPY targets raises transferred MLP accuracy from 0.586 to 0.597, and the
within-cohort model still performs better. Finally, an earlier 16-class
Austrian crop pilot with 128 test polygons showed some utility for converted
v1. However, even on true v2 inputs the head reached only about 0.20 accuracy, so
we treat this pilot as supplementary.

\section{Discussion and conclusion}
\label{sec:discussion}

The frozen-head tests show substantial failure when raw inputs come from
another release, even though the vector dimensions match. Translation recovers
much of the target head's performance on the tested sites. On the main Esri
task, converted v1.1 trails true v2 by 0.020 accuracy and 0.049 macro-F1.
Individual predictions can still differ, and the site-held-out results do not
establish a fixed non-inferiority margin across regions. The compatibility we
observe applies to the specified releases, v2 target source, sites, year,
labels, and head.

Direct v1$\rightarrow$v2 translation outperforms the composition of adjacent
converters on the same test vectors. On the Esri task, direct converted v1
reaches 0.808 macro-F1, compared with 0.749 after composition. The second map
may amplify the errors of the first, or v1.1 may omit information useful for
the final target, and our experiment does not separate these possibilities.
In this setting, an intermediate release adds error rather than providing a
lossless route to v2.

On new sites, the original converter reconstructs v2 embeddings more closely
on average than the within-cohort converter, but its predictions through the
frozen head are worse. Because MSE weights every embedding dimension equally,
a small shift across a classifier boundary could account for this mismatch,
although we have not tested that mechanism directly. Work on neural network
representations points the same way: model stitching can reveal differences
between representations that similarity measures such as centred kernel
alignment cannot \cite{bansal2021stitching,kornblith2019cka}. The transfer
gap is also an instance of the domain shift long studied in remote sensing
classification \cite{tuia2016domain}. Our estimate of it is limited, because
the new-site Esri test is dominated by class 2 and has only four independent
test sites. WorldCover also has four test sites and substantial site-to-site
variation. More geographic holdouts would be needed to estimate performance
over a deployment area.

The 24-site pilot permits pseudo-v2 inference at sampled points where the
queried v2 product returns no valid vector. Native v2 is unavailable at the
same locations, so the pilot cannot assess reconstruction fidelity or decision
agreement with a native-v2 head, nor does it characterize coverage in
surrounding tiles or other years. Two further limitations apply throughout.
First, the Esri and WorldCover labels are themselves map products with
classification errors \cite{venter2022landcover}. Second, stored and freshly
read v2 targets differ numerically for an unknown reason. These limitations
should be reported alongside the converter and head versions.

A small local adaptation experiment could test whether a few paired samples
reduce the geographic transfer gap. Its calibration pairs would need to be
disjoint from the local evaluation points, with the head kept frozen. That
experiment remains to be run. Meanwhile, any use of pseudo-v2 embeddings
should record target provenance and inspect performance by region and class.

In summary, raw embeddings from different Tessera releases are not
interchangeable, but a converter fitted on a few thousand co-located pairs
can keep an existing classifier usable on the tested sites. Our results
suggest three practices for users who adopt a new release: fit the converter
directly between the two releases of interest, fit it on data near the
intended deployment area, and validate it through the downstream classifier
rather than through embedding similarity.

\section*{Acknowledgment}
OpenAI Codex assisted with drafting and editing the abstract and Sections I--IV, and with figure preparation. The authors are responsible for the scientific content and final manuscript.
\bibliographystyle{IEEEtran}
\bibliography{references}
\end{document}